\documentclass[10pt,twocolumn,letterpaper]{article}

\usepackage{iccv}
\usepackage{times}
\usepackage{epsfig}
\usepackage{graphicx}
\usepackage{amsmath}
\usepackage{amssymb}
\usepackage[accsupp]{axessibility}

\usepackage{booktabs}
\usepackage{multirow}
\usepackage{wrapfig}
\usepackage{xcolor}

\usepackage{listings}
\newcommand{\E}{\mathcal{E}}
\newcommand{\F}{\mathcal{F}}
\newcommand{\G}{\mathcal{G}}

\newcommand{\V}{\mathcal{V}}

\newcommand{\h}{\mathbf{h}}

\newcommand{\splitatcommas}[1]{%
  \begingroup
  \begingroup\lccode`~=`, \lowercase{\endgroup
    \edef~{\mathchar\the\mathcode`, \penalty0 \noexpand\hspace{0pt plus 1em}}%
  }\mathcode`,="8000 #1%
  \endgroup
}

\usepackage[pagebackref=true,breaklinks=true,colorlinks,bookmarks=false]{hyperref}

\usepackage[capitalize]{cleveref}
\Crefname{figure}{Figure}{Figures}
\Crefname{listing}{Program}{Programs}

\iccvfinalcopy %

\begin{document}

\title{PlankAssembly: Robust 3D Reconstruction from Three Orthographic Views with Learnt Shape Programs}

\author{
Wentao Hu\textsuperscript{1,2}$^{\dagger*}$ \quad
Jia Zheng\textsuperscript{3}$^*$ \quad
Zixin Zhang\textsuperscript{4}$^*$ \quad
Xiaojun Yuan\textsuperscript{4} \quad
Jian Yin\textsuperscript{1,2} \quad
Zihan Zhou\textsuperscript{3} \\
\textsuperscript{1}Sun Yat-Sen University \quad
\textsuperscript{2}Guangdong Key Laboratory of Big Data Analysis and Processing \\
\textsuperscript{3}Manycore Tech Inc. \quad
\textsuperscript{4}University of Electronic Science and Technology of China \\
{\tt\small \url{https://manycore-research.github.io/PlankAssembly}}
}

\maketitle

\begin{abstract}
In this paper, we develop a new method to automatically convert 2D line drawings from three orthographic views into 3D CAD models. Existing methods for this problem reconstruct 3D models by back-projecting the 2D observations into 3D space while maintaining explicit correspondence between the input and output. Such methods are sensitive to errors and noises in the input, thus often fail in practice where the input drawings created by human designers are imperfect. To overcome this difficulty, we leverage the attention mechanism in a Transformer-based sequence generation model to learn flexible mappings between the input and output. Further, we design shape programs which are suitable for generating the objects of interest to boost the reconstruction accuracy and facilitate CAD modeling applications. Experiments on a new benchmark dataset show that our method significantly outperforms existing ones when the inputs are noisy or incomplete.
\end{abstract}

\newcommand\blfootnote[1]{
\begingroup
\renewcommand\thefootnote{}\footnote{#1}
\addtocounter{footnote}{-1}
\endgroup
}
\blfootnote{$^\dagger$Work done during internship at Manycore Tech Inc.}
\blfootnote{$^*$Equal contributions.}

\section{Introduction}

In this paper, we tackle a long-standing problem in computer-aided design (CAD), namely 3D object reconstruction from three orthographic views. In today's product design and manufacturing industry, 2D engineering drawings are commonly used by designers to realize, update, and share their ideas, especially during the initial design stages. But to enable further analysis (\eg, finite element analysis) and manufacturing, these 2D designs must be manually realized as 3D models in CAD software. Therefore, if a method can automatically convert the 2D drawings into 3D models, it would greatly facilitate the design process and improve overall efficiency.

\begin{figure}[t]
  \centering
  \setlength{\tabcolsep}{2pt}
  \begin{tabular}{ccc}
    Front & Top & Side
    \tabularnewline
    \includegraphics[height=0.75in]{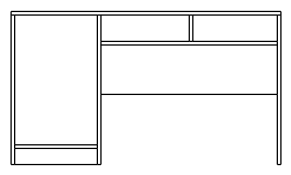} & 
    \includegraphics[height=0.75in]{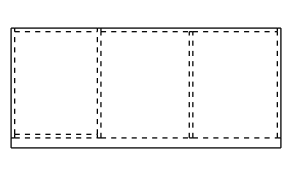} & 
    \includegraphics[height=0.75in]{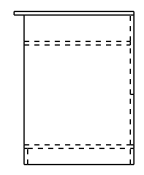} 
    \tabularnewline
    \includegraphics[height=0.75in]{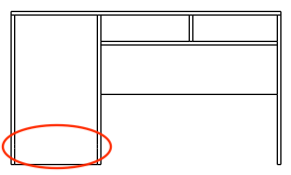} & 
    \includegraphics[height=0.75in]{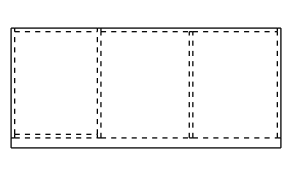} & 
    \includegraphics[height=0.75in]{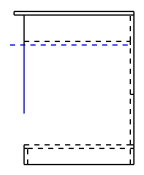} 
    \tabularnewline
    \includegraphics[height=0.75in]{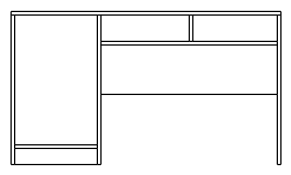} & 
    \includegraphics[height=0.75in]{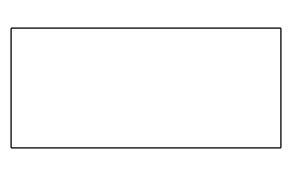} & 
    \includegraphics[height=0.75in]{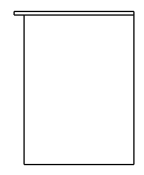} 
    \tabularnewline
  \end{tabular}
  \caption{An illustration of various input drawings. {\bf From top to bottom}: clean inputs, noisy inputs, and visible inputs. We use solid and dashed lines to represent the visible and hidden lines, respectively. In noisy line drawings (the second row), we use blue lines to represent noisy lines and highlight the missing lines using the red circle.}
  \label{fig:teaser}
\end{figure}

As the most popular way to describe an object in 2D drawings, an orthographic view is the projection of the object onto the plane that is perpendicular to one of the three principal axes (\cref{fig:teaser}). Over the past few decades, 3D reconstruction from three orthographic views has been extensively studied, with significant improvements in terms of the types of applicable objects and computational efficiency~\cite{SakuraiG83, GuTS86, LequetteR88, YanCT94, YouY96, ShinS98, Kuo98, LiuHCS01, GongZZS06a, GongZZS06b}. However, to the best of our knowledge, these techniques have not enjoyed wide adoption in CAD software and commercial products.

Among the challenges faced by existing methods in practice, their sensitivity to errors and missing components in the drawings is arguably the most critical one. To understand this issue, we note that almost all existing methods follow a standard procedure for 3D reconstruction, which consists of the following steps: {\bf (i)} generate 3D vertices from 2D vertices; {\bf (ii)} generate 3D edges from 3D vertices; {\bf (iii)} generate 3D faces from 3D edges; and {\bf (iv)} construct 3D models from 3D faces (see \cref{fig:noise} for an illustration). One main benefit of following the pipeline is that all solutions that match the input views can be found, as it establishes explicit correspondences between entities in the 3D model and those in the drawing. But in practice, rather than making an extra effort to perfect the drawings, designers would deem a drawing good enough as long as it conveys their ideas. Hence, some entities may be erroneous or missing. As a result, the aforementioned pipeline often fails to find the desired solution.

To overcome this difficulty, therefore, it is necessary to reason about the 2D drawings in a more holistic manner, and enable more flexible mappings between the input and output. Recently, Transformer~\cite{VaswaniSPUJGKP17} has become the standard architecture in many NLP and CV tasks. It is particularly effective in sequence-to-sequence (seq2seq) problems, such as machine translation, where reasoning about the context and soft alignment between the input and output are critical. Motivated by this, we convert our problem into a seq2seq problem and propose a Transformer-based deep learning method. Intuitively, the self-attention modules allow the model to capture the intent of the product designers even if their drawings are imperfect, and the cross-attention modules enable flexible mappings between geometric entities in the 2D drawing and 3D model.

\begin{figure}[t]
  \centering
  \includegraphics[width=0.8\linewidth]{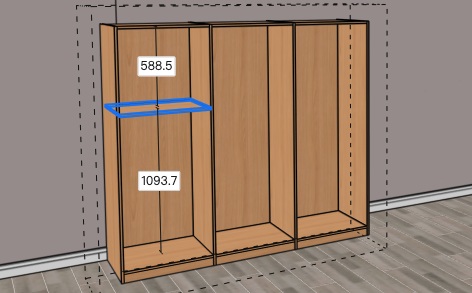}
  \caption{An illustration of cabinet design in a 3D modeling software.}
  \label{fig:cabinet}
\end{figure}

Another benefit of employing learned representations and soft alignments for geometric entities is that one is free to choose how the 3D model is constructed. This provides us with opportunities to incorporate domain knowledge in our method to boost its performance. To illustrate this, we focus on a specific type of product, \emph{cabinet furniture}, in this paper. As illustrated in \cref{fig:cabinet}, a cabinet is typically built by arranging and attaching a number of planks (\ie, wooden boards) together in a 3D modeling software. 
To this end, we develop a simple domain-specific language (DSL) based around declaring planks and then attaching them to one another, so that each cabinet can be represented by a program. Finally, given the input orthographic views, we train the Transformer-based model to predict the program associated with the cabinet.

To systematically evaluate the methods, we build a new benchmark dataset consisting of more than 26,000 3D cabinet models for this task. Most of them are created by professional interior designers using commercial 3D modeling software. Extensive experiments show that our method is much more robust to imperfect inputs. For example, the traditional method achieves an F1 score of $8.20\%$ when $30\%$ of the lines are corrupted or missing in the input drawings, whereas our method achieves an F1 score of $90.14\%$.

In summary, the contributions of this work are: {\bf (i)} To the best of our knowledge, we are the first to use deep generative models in the task of 3D CAD model reconstruction from three orthographic views. Compared to existing methods, our model learns a more flexible mapping between the input and output, thus being more robust to noisy or incomplete inputs. {\bf (ii)} We propose a new network design that learns shape programs to assemble planks into 3D cabinet models. Such a design not only improves reconstruction accuracy but also facilitates downstream applications such as CAD model editing.

\section{Related Work}

\noindent \textbf{3D reconstruction from three orthographic views.} Studies on recovering 3D models from three orthographic views date back to the 70s and 80s~\cite{IdesawaM73, MarkowskyW80, WesleyM81, SakuraiG83, GuTS86}. An early survey on this topic appears in~\cite{WangG93}. According to~\cite{WangG93}, to obtain 3D objects in the boundary representation (B-rep) format, existing methods follow a four-stage scheme in which 3D vertices, edges, faces, and blocks are gradually built upon the results of previous steps. As mentioned before, a key strength of the framework is that all possible solutions that exactly match the input views can be found.

Subsequent methods for this task~\cite{YanCT94, YouY96, ShinS98, Kuo98, LiuHCS01, GongZZS06a, GongZZS06b} also follow the same procedure and focus on extending the methods' applicable domain to cover more types of objects. For example, Shin and Shin~\cite{ShinS98} developed a method to reconstruct objects composed of planar and limited quadric faces, such as cylinders and tori, that are parallel to one of the principal axes. To remove the restriction placed on the axes of curved surfaces, Liu~\etal~\cite{LiuHCS01} designed an algorithm that combines the geometric properties of conics with affine properties. Later, Gong~\etal~\cite{GongZZS06b} proposed to recognize quadric surface features via hint-based pattern matching in the Link-Relation Graph (LRG), expanding the applicable domain to cover objects like those with interacting quadric surfaces. However, all these methods assume clean inputs and, as we will show in the experiment section, could easily break down in the presence of errors and noises. 

\begin{figure*}[t]
  \centering
  \includegraphics[width=0.99\linewidth]{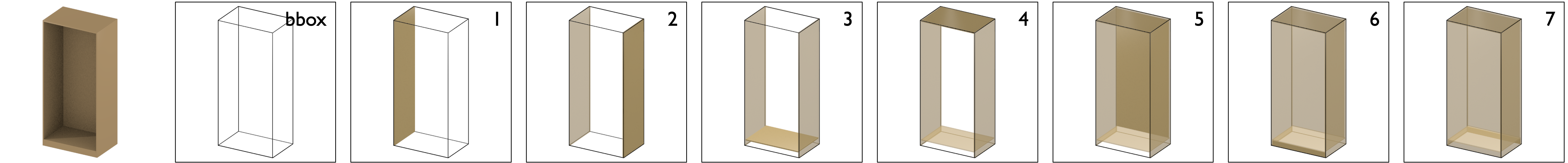}
  \caption{An illustration of how a simple cabinet is incrementally constructed by executing the shape program commands. The corresponding shape program is shown in \cref{program}.}
  \label{fig:steps}
\end{figure*}

Recently, Han~\etal~\cite{HanXLWF20} also trained a deep network to reconstruct a 3D model from three orthographic views. However, their method takes raster images as input and produces results in the format of unstructured point clouds, which are of little use in CAD modeling applications. In contrast, our method directly uses vectorized line drawings as input and generates structured CAD models as output.

\smallskip
\noindent \textbf{Deep generative models for CAD.} With the availability of large-scale CAD datasets such as ABC~\cite{KochMJWABAZP19} and Fusion 360 Gallery~\cite{WillisPLCDLSM21}, a line of recent work trains deep networks to generate structured CAD data in the form of 2D sketches~\cite{WillisJLCP21, GaninBLKS21, SeffZRA22} or 3D models~\cite{WuXZ21, JayaramanLDWSM23, GuoLPLTG22, XuWLCJF22}. These methods all cast it as a sequence generation problem, but differ in the DSLs used to produce the output. Our idea to generate cabinet furniture by assembling plank models together is inspired by ShapeAssembly~\cite{JonesBX0JGMR20}, which learns to generate objects as hierarchical 3D part graphs. However, unlike the above studies, which focus on the generative models themselves, we propose to use generative models to build effective and efficient method for 3D CAD model reconstruction from three orthographic views.

\section{A Simple Assembly Language for Cabinets}
\label{sec:formulation}

In this section, our goal is to define a domain-specific language (DSL) for the shapes of interest (\ie, cabinet furniture). With this language, each cabinet model can be represented by a shape program, which will later be converted into a sequence of tokens as the output of a Transformer-based seq2seq model.

We define our DSL in the way that the resulting shape program resembles how a human designer builds the model in 3D modeling software. As shown in \cref{fig:cabinet}, a cabinet is typically assembled by a list of plank models. In practice, most planks are axis-aligned cuboids. Therefore, we use cuboid as the only data type in our language. In \cref{sec:discussion}, we discuss how our approach may be extended to accommodate more complex shapes (\eg, a plank with a non-rectangular profile).

An axis-aligned cuboid has six degrees of freedom (DOF), which correspond to the starting and ending coordinates along the three axes:
\begin{equation}
  \texttt{Cuboid}(x_{\min}, y_{\min}, z_{\min}, x_{\max}, y_{\max}, z_{\max}).
  \label{eq:cuboid}
\end{equation}
In practice, instead of specifying the numerical values for all the coordinates, human designers frequently use the \emph{attachment} operation. As a form of geometric constraints, the benefit of using attachment is at least two-fold: \emph{First}, it enables users to quickly specify the location of a plank without explicitly calculating (some of) its coordinates; \emph{Second}, it facilitates future edits as any changes made to a plank will be automatically propagated to the others. Take \cref{fig:cabinet} as an example. When adding a plank (highlighted in blue), a designer may attach its four sidefaces to existing planks (including the invisible bounding box), while specifying the distances to the top and bottom in numerical values.

\begin{wrapfigure}{r}{0.3\linewidth}
\begin{lstlisting}[language=C++,basicstyle=\small\ttfamily,numbers=none]
union Coord{
  float v;
  float* p;
};
\end{lstlisting}
\end{wrapfigure}
Our language supports specifying the plank coordinates via either numerical values or attachment operation by adopting the $\texttt{Union}$ structure commonly used in programming languages (\eg, C++). As shown in the figure to the right, each of the six coordinates in \cref{eq:cuboid} can either take a numerical value or be a pointer to the corresponding coordinate of another cuboid (to which it attaches to). \cref{fig:steps} shows an example cabinet incrementally constructed by imperatively executing the program commands (\cref{program}).

\smallskip
\noindent {\em Shape program as a DAG.} Alternatively, we may interpret the shape program as a directed acyclic graph (DAG). Note that each plank model consists of six faces, where each face corresponds to exactly one DOF in the axis-aligned cuboid (\ie, $x_{\min}, y_{\min}, z_{\min}, x_{\max}, y_{\max}, z_{\max}$). Therefore, each program can be characterized by a graph $\G = \{ \F, \E \}$, whose vertices $\F = \{ f_1, \ldots, f_{|\F|} \}$ represent the faces of plank models and whose edges $\E = \{ e_1, \ldots, e_{|\E|} \}$ represent attachment relationships between faces. Each directed edge $e_{i \rightarrow j}$ is an ordered pair of vertices $(f_i, f_j)$, indicating the $i$-th face $f_i$ attaches to the $j$-th face $f_j$. We assume that each face can attach to at most one another face; that is, the out-degree of any face $f_i$ is at most one. Further, the edges $\E$ can be represented by an adjacency matrix $A \in \mathbb{R}^{|\F| \times |\F|}$. Specifically, $A_{ij}$ is 1 if $f_i$ directs to $f_j$, and 0 otherwise.

\begin{lstlisting}[float=tp,belowskip=-2em,caption={The subscript indicates the index of coordinate values.},label=program]
bbox = Cuboid(-0.35, -0.23, -0.76, 0.35, 0.23, 0.76)
plank1 = Cuboid(bbox$\textsubscript{1}$, bbox$\textsubscript{2}$, bbox$\textsubscript{3}$, -0.34, bbox$\textsubscript{5}$, bbox$\textsubscript{6}$)
plank2 = Cuboid(0.34, bbox$\textsubscript{2}$, bbox$\textsubscript{3}$, bbox$\textsubscript{4}$, bbox$\textsubscript{5}$, bbox$\textsubscript{6}$)
plank3 = Cuboid(plank1$\textsubscript{4}$, bbox$\textsubscript{2}$, -0.70, plank2$\textsubscript{1}$, bbox$\textsubscript{5}$, -0.69)
plank4 = Cuboid(plank1$\textsubscript{4}$, bbox$\textsubscript{2}$, 0.75, plank2$\textsubscript{1}$, bbox$\textsubscript{5}$, bbox$\textsubscript{6}$)
plank5 = Cuboid(plank1$\textsubscript{4}$, 0.21, plank3$\textsubscript{6}$, plank2$\textsubscript{1}$, 0.22, plank4$\textsubscript{3}$)
plank6 = Cuboid(plank1$\textsubscript{4}$, bbox$\textsubscript{2}$, bbox$\textsubscript{3}$, plank2$\textsubscript{1}$, -0.21, plank3$\textsubscript{3}$)
plank7 = Cuboid(plank1$\textsubscript{4}$, 0.21, bbox$\textsubscript{3}$, plank2$\textsubscript{1}$, bbox$\textsubscript{5}$, plank3$\textsubscript{3}$)
\end{lstlisting}

\section{The PlankAssembly Model}

As shown in \cref{fig:teaser}, we assume the input consists of three orthographic projections of the object, namely, the front view, top view, and side view: $\V = \{V_F, V_T, V_S\}$. Each view can be regarded as a planar graph of 2D edges and node points where the edges meet. We use solid lines to represent visible edges and dashed lines to represent hidden edges. Our goal is to reconstruct a 3D cabinet model described by the shape program or the equivalent DAG $\G$.

In this paper, we cast 3D reconstruction as a seq2seq problem. In \cref{sec:method:input,sec:method:output}, we describe how to encode the input views $\V$ and the shape program $\G$ as 1D sequences $\V^\text{seq}$ and $\G^\text{seq}$, respectively. Then, we introduce the design of our PlankAssembly model, which adopts a Transformer-based encoder-decoder architecture to learn the probability distribution $p (\G^\text{seq} \mid \V^\text{seq})$, in \cref{sec:method:network}. Finally, we present implementation details in \cref{sec:method:train}.

\subsection{Input Sequences and Embeddings}
\label{sec:method:input}

For the input conditions, we first order the 2D edges in $\V$ by the views. Each 2D edge is written as $(x_1, y_1, x_2, y_2)$, where we order its two endpoints from lowest to highest by the $x$-coordinate, followed by the $y$-coordinate (if $x_1 = x_2$). Then, we order a set of 2D edges by $x_1$, followed by $x_2$, $y_1$, and $y_2$. Next, we flatten all the edges into a 1D sequence $\V^\text{seq} = \{v_1, \ldots, v_{N_{v}}\}$. Note that, since each 2D edge has four DOFs (\ie, the $x$- and $y$-coordinates of two endpoints), the length of $\V^\text{seq}$ is $N_{v} = 4N_\text{edge}$, where $N_\text{edge}$ is the total number of 2D edges in all three orthographic views.

We embed the $i$-th token $v_i$ as:
\begin{multline}
  E (v_i) = E_\text{value} (v_i) + E_\text{view} (v_i) + E_\text{edge} (v_i) \\
  + E_\text{coord} (v_i) + E_\text{type} (v_i),
  \label{eq:input-embed}
\end{multline}
where the value embedding $E_\text{value}$ indicates the quantized coordinate value of the token, the view embedding $E_\text{view}$ indicates which view (\ie, the front, top, or side view) the 2D edge is from, the edge embedding $E_\text{edge}$ indicates the relative position of the 2D edge in the corresponding view, and the coordinate embedding $E_\text{coord}$ indicates the relative position of the coordinate in the corresponding 2D edge. Finally, we use a type embedding $E_\text{type}$ to indicate whether the 2D edge is visible or hidden. In this paper, we quantize the coordinate values into 9-bit integers and use learned 512-D embeddings for each term in \cref{eq:input-embed}.

\subsection{Output Sequences and Embeddings}
\label{sec:method:output}

To generate the shape program sequentially, we need to map the graph $\G$ to a sequence $\G^\text{seq}$. This requires us to define a vertex order $\pi$ on $\G$: We first sort the vertices topologically, ensuring that direct successors are listed before their corresponding direct predecessors. Then, vertices that are not directly connected are ordered by the coordinate values. This gives us a sorted graph $\G^\pi$ whose vertices $\F^\pi$ follow the order $\pi$.

Since we would like to capture the modeling process and facilitate future editing, we prioritize attachment relationships over geometric entities. Similar to the input sequence encoding, we flatten $\G^\pi$ to obtain a 1D sequence $\G^\text{seq}$. The $i$-th element of the sequence $\G^\text{seq}$ can be obtained as:
\begin{align}
  g_i = 
  \begin{cases}
    f^{\pi}_i, & \text{if }A^{\pi}_{ij} = 0, \forall j, \\
    e^{\pi}_{i \rightarrow j}, & \text{if }A^{\pi}_{ij} = 1.
  \end{cases}
\end{align}%
Further, we use two special tokens, \texttt{[SOS]} and \texttt{[EOS]}, to indicate the start and end of the output sequence, respectively.

For the inputs to the decoder of our model, we embed the token $g_i$ using the associated face $f^{\pi}_i$ as follows:
\begin{align}
  E(g_i) = E(f^{\pi}_i) = E_\text{value} (f^{\pi}_i) + E_\text{plank} (f^{\pi}_i) + E_\text{face} (f^{\pi}_i).
\end{align}%
The value embedding $E_\text{value}$ indicates the quantized coordinate value, which is shared for the input and output sequences. The plank embedding $E_\text{plank}$ indicates the location of the corresponding plank in the cabinet model, and the face embedding $E_\text{face}$ indicates the relative position of the face within the plank.

If the token corresponds to an edge $e^{\pi}_{i \rightarrow j}$ in $\G$, we identify the face $f^{\pi}_j$ to which the current face $f^{\pi}_i$ attaches, and use the same value embedding as $f^{\pi}_j$.

\subsection{Model Design} 
\label{sec:method:network}

To tackle this seq2seq problem, we factorize the joint distribution over the output sequence into a series of conditional distributions:
\begin{align}
  p (\G^\text{seq} \mid \V^\text{seq}) = \prod_t p \left( g_t \mid \G_{<t}^\text{seq}, \V^\text{seq} \right).
\end{align}
Here, since $g_t$ may take the form of either a geometric entity (\ie, $f^{\pi}_i$) or an attachment relationship (\ie, $e^{\pi}_{i \rightarrow j}$), we need to generate a probability distribution over a fixed-length vocabulary set (of quantized coordinate values) \emph{plus} a variable-length set of tokens $\G_{<t}^\text{seq}$ in the output sequence.

The former distribution is a categorical distribution, which is commonly used in classification tasks. Let $\h_t$ be the hidden feature obtained by the decoder at time $t$, we project it to the size of the vocabulary via a linear layer, which is then normalized to form a valid distribution:
\begin{align}
  p_{\text{vocab}}(g_t \mid \G_{<t}^\text{seq}, \V^\text{seq}) = \operatorname{softmax} \left( \operatorname{linear} \left( \h_t \right) \right).
\end{align}

To generate a distribution over the output sequence $\G^\text{seq}_{<t}$ at time $t$, we adopt the Pointer Networks~\cite{VinyalsFJ15}. Specifically, we first use a linear layer to predict a pointer. The pointer is then compared with the hidden features of all former steps via dot-product. Finally, a distribution over the output sequence is obtained via a softmax layer:
\begin{multline}
  p_{\text{attach}}(g_t\rightarrow g_k \mid \G_{<t}^\text{seq}, \V^\text{seq}) = \\ \operatorname{softmax}_k \left( \operatorname{linear} \left( \h_t \right)^T \h_{<t} \right).
\end{multline}

Instead of directly comparing these two distributions, we follow Pointer-Generator Networks~\cite{SeeLM17} and introduce an attachment probability $w_t$ to weight these two distributions. The attachment probability $w_t$ is obtained via a linear layer and a sigmoid function $\sigma(\cdot)$: $w_t = \sigma \left( \operatorname{linear} \left( \h_t \right) \right)$. Thus, the final distribution is the concatenation of the two weighted distributions:
\begin{align}
  p (g_t \mid \G_{<t}^\text{seq}, \V^\text{seq}) = \operatorname{concat}\big\{(1 - w_t)\cdot p_{\text{vocab}},  w_t\cdot p _{\text{attach}} \big\}.
  \label{eq:probability}
\end{align}

Finally, given a training set, the parameters of the model can be learned by maximizing the conditional distributions \cref{eq:probability} via a standard cross-entropy loss.

\begin{figure}[t]
  \centering
  \includegraphics[width=0.9\linewidth]{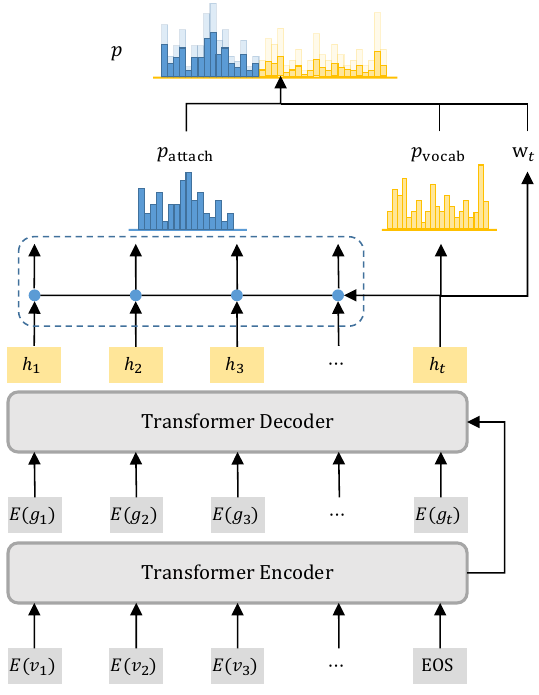}
  \caption{Network architecture. Our model takes the line sequences as input and outputs the shape program sequence auto-regressively. At each time step $t$, the Transformer decoder outputs an attachment distribution $p_\text{attach}$ over the previously predicted outputs, a vocabulary distribution $p_\text{vocab}$, and an attachment probability $w_t$. The final distribution $p$ is obtained by concatenation of the two weighted distributions.}
  \label{fig:network}
\end{figure}

\smallskip
\noindent \textbf{Network architecture.} We use the standard Transformer blocks~\cite{VaswaniSPUJGKP17} as the basic blocks of our PlankAssembly model. Given the input embeddings $\{E(v_1), E(v_2), \ldots\}$, the encoder encodes them into contextual embeddings. At decoding time $t$, the decoder produces hidden feature $\h_t$ based on the contextual embeddings and the decoder inputs $\{E(g_1), E(g_2), \ldots\}$. We use 6 Transformer layers for both the encoder and the decoder. Each layer has a feed-forward dimension of 1024 and 8 attention heads. The network architecture is summarized in \cref{fig:network}.

\subsection{Implementation Details}
\label{sec:method:train}

\noindent \textbf{Training.} We implement our models with PyTorch Lightning~\cite{lightning}.  We use 6 Transformer layers for both the encoder and the decoder. Each layer has a feed-forward dimension of 1024 and 8 attention heads. The network is trained for 400K iterations on four NVIDIA RTX 3090 GPU devices. We use Adam optimizer~\cite{KingmaB15} with a learning rate of $10^{-4}$. The batch size is set to 16 per GPU.

\smallskip
\noindent \textbf{Inference.} At inference time, we take several steps to ensure valid predictions from our model. First, we observed that two attaching faces must correspond to the opposite DOFs on the same axis, in order to avoid any spatial conflicts. For example, the $x_{\min}$ token of one plank can only point to the $x_{\max}$ token of another plank, and vice versa. Thus, we mask all invalid positions during inference. Second, we filter out the predicted planks with zero volume.

\section{Experiments}

\subsection{Experimental Setup}

\noindent \textbf{Dataset.} We create a large-scale benchmark dataset for this task, taking advantage of access to a large repository of cabinet furniture models from Kujiale\footnote{\url{http://kujiale.com}}, an online 3D modeling platform in the interior design industry. Most models in the repository are created by professional designers using commercial parametric modeling software, and are used for real-world production.

Several rules are used to filter the data: (i) We remove duplicated 3D models based on the similarity of the three orthographic views; (ii) We exclude models with fewer than four planks, more than 20 planks, or more than 300 edges in total. The remaining data is randomly split into three parts: 24039 for training, 1329 for validation, and 1339 for testing. To synthesize the three orthographic views, we use the \texttt{HLRBRep\_Algo} API from pythonOCC~\cite{pythonocc}, which is built upon the Open CASCADE Technology modeling kernel~\cite{occt}.

For our task, we need to parse each parametric cabinet model into a shape program. We first obtain the planks by extracting the geometric entities in the cabinet model. Note that in the parametric modeling software, a plank is typically created by first drawing a 2D profile and then applying the extrusion command. Thus, we categorize the faces of each plank into \textit{sideface} or \textit{endface}, depending on whether they are along the direction of the extrusion or not. Then, given a pair of faces from two different planks, we consider that an attachment relationship exists if (i) the two faces are within a distance threshold of 1mm, and (ii) the pair consists of one sideface and one endface. Finally, a directed edge from the endface to the sideface is added in $\G$.

\smallskip
\noindent \textbf{Evaluation metrics.} To evaluate the quality of the 3D reconstruction results, we use three standard metrics: precision, recall, and F1 score. Specifically, for a cabinet model, we use Hungarian matching to match the predicted planks and the ground truth planks. A prediction is considered a true positive if its 3D intersection-over-union (IOU) with one ground truth is greater than $0.5$.

\subsection{Comparison to Traditional Methods}

In this section, we systematically compare our approach with the traditional methods for 3D reconstruction from three orthographic views. Since no implementation of the traditional pipeline is publicly available, we reimplement the pipeline by closely following prior work~\cite{SakuraiG83, ShinS98}. Recall that, starting from the input views, the traditional pipeline generates 3D vertices, 3D edges, 3D faces, and 3D blocks step by step. Then, solutions are found by enumerating all combinations of the candidate blocks and checking if their 2D projections match the input views. 

To make the pipeline suitable for reconstructing assembly models like cabinets, we introduce two minor adjustments to it. \emph{First}, in the traditional pipeline, two blocks that share a common face are merged together, which leads to mismatching between projections and input line drawings. In our implementation, we simply omit the original merging operation. \emph{Second}, the blocks generated by the traditional pipeline correspond to the minimal closed spaces in 3D. During the evaluation, they cannot be directly matched with ground truth planks through bipartite matching. Instead, we group the blocks as a single prediction if they overlap with the same ground-truth plank model. Note that the proposed adjustments slightly favor the traditional approach, as we use ground truth information to merge the blocks.

Moreover, in cases where the traditional approach generates multiple solutions that all satisfy the inputs, we randomly select one as the final output.

\begin{figure}[t]
  \centering
  \includegraphics[width=0.9\linewidth]{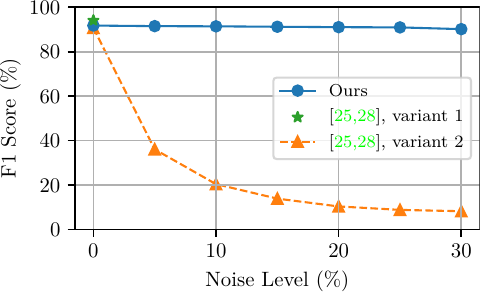}
  \caption{Comparison on varying input noise levels.}
  \label{fig:results}
\end{figure}

\smallskip
\noindent{\bf Experiment on varying input noise levels.} We first study the performance of both methods on imperfect inputs. In this experiment, we inject varying levels of noise into the input views. We consider two types of noises/errors commonly seen in real-world drawings: missing lines and inaccurate endpoints. Specially, we randomly select a percentage of 2D edges in the input views. For each selected edge, we either delete it or randomly perturb its endpoints along the edge direction. \cref{fig:results} reports the F1 scores of both methods as the percentage varies from $0\%$ to $30\%$.

\begin{figure*}[t]
  \centering
  \includegraphics[width=0.9\linewidth]{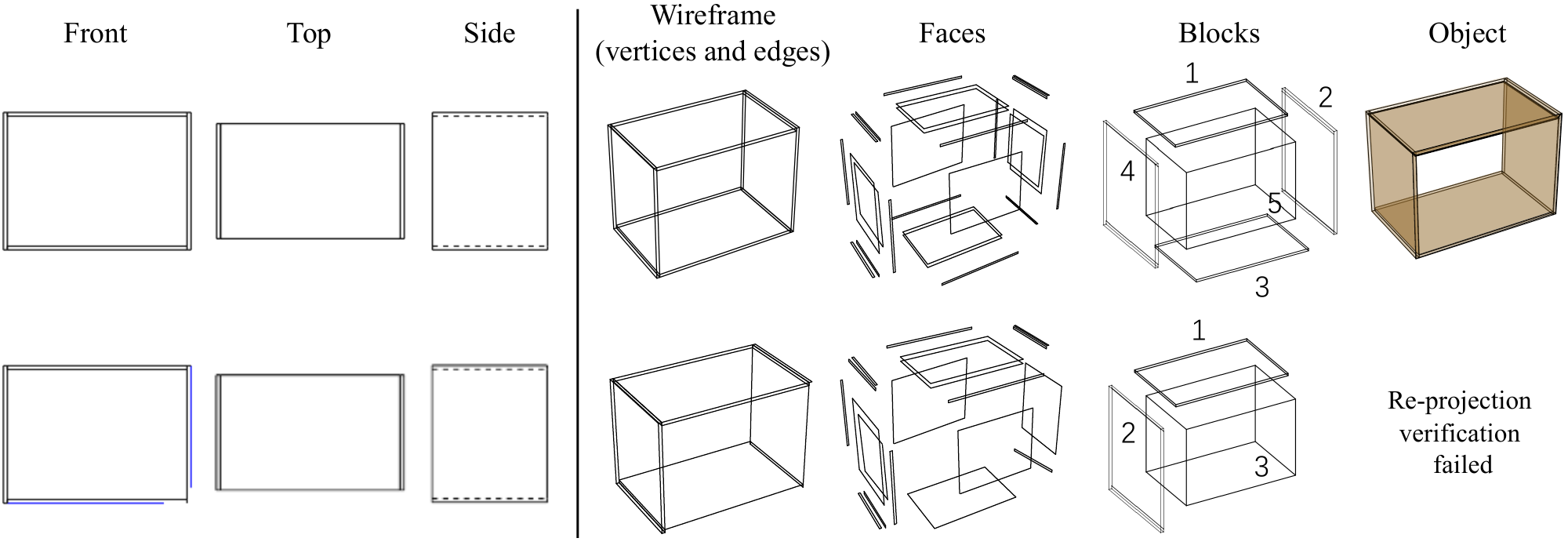}
  \caption{A step-by-step illustration of the traditional approach on clean input {\bf (top)} and noisy input {\bf (bottom)}.  The traditional pipeline accurately reconstructs the object when the input line drawings are free of noise. However, it fails to recover all the 3D blocks in the presence of noises, resulting in a failure of the re-projection verification in the final solid reconstruction step.}  
  \label{fig:noise}
\end{figure*}

Note that, when applied to our dataset, a major bottleneck of the traditional pipeline is the re-projection verification step in which all possible combinations of candidate blocks are projected to 2D to check if they match the input views. The reason is two-fold. \emph{First}, such a match typically does not exist for noisy inputs. This is illustrated in \cref{fig:noise}. When the input views contain errors, the 3D blocks generated by the traditional pipeline are often incomplete. Consequently, none of the combinations would exactly match the input views, resulting in a failure in the final solid reconstruction step. \emph{Second}, even on clean inputs, the verification may take a long time due to a large number of possible combinations of the candidate blocks.

Thus, for completeness, we compare two variants of traditional pipeline in \cref{fig:results}. In \emph{the first variant}, we enforce the re-projection verification step during reconstruction. On clean inputs, this variant achieves a slightly higher F1 score ($94.07\%$) than ours ($91.75\%$). However, this variant fails to produce a solution for 518 cases (out of 1339 test cases) in a reasonable time (5 minutes) due to exponential search complexity. Furthermore, this variant is not applicable to noisy inputs.

In \emph{the second variant}, we ignore the re-projection verification step and directly use the union of blocks as the solution. As shown in \cref{fig:results}, it achieves an F1 score of $90.67\%$ on clean input. And its performance degrades quickly as the input noise level increases. On the $30\%$ noise level, this variant only produces results on 54 objects and has an F1 score of $8.20\%$. 

In contrast, our approach is much more robust to input noises. Specifically, its performance only slightly drops from $91.75\%$ to $90.14\%$ as the noise level increases from $0\%$ to $30\%$, verifying the key advantage of our method over traditional ones. In terms of inference time, our method takes about $0.63$ seconds per sample on a single RTX 3090 GPU device.

\smallskip
\noindent\textbf{Experiment on inputs with visible edges only.} In real-world design practice, it is common for designers to omit the hidden edges of line drawings. Although it is still easy for humans to infer the 3D model, the traditional approach is likely to fail since the inputs are highly incomplete. To further demonstrate the robustness of our method, we conduct an experiment in which only the visible parts of the line drawings as used as input. For this experiment, we remove all invisible edges in the training set and follow the same protocol in \cref{sec:method:train} to train our network from scratch.

\begin{table}[t]
  \centering
  \begin{tabular}{l|ccc}
    \toprule
    Methods & Precision & Recall & F1 score \tabularnewline
    \midrule
    \cite{SakuraiG83, ShinS98} & {\bf 99.64} & 26.47 & 39.31 \tabularnewline
    Ours & 84.12 & {\bf 82.05} & {\bf 82.62} \tabularnewline
    \bottomrule
  \end{tabular}
  \caption{Comparison on inputs with visible edges only.}
  \vspace{-3mm}
  \label{tab:visible}
\end{table}

As shown in \cref{tab:visible}, the traditional approach performs poorly on this task, with very low recall and F1 score. This is expected because, on average, invisible edges account for about $48\%$ of the edges in a line drawing. Meanwhile, our method is robust to the incomplete inputs, achieving an F1 score of $82.62\%$.

\begin{figure*}[t]
  \centering
  \setlength{\tabcolsep}{1pt}
  \setlength{\fboxsep}{0pt}
  \renewcommand{\arraystretch}{0}
  \begin{tabular}{c|c|c|c|c||c|c|c|c|c}
    \toprule
    Front & Top & Side & \cite{SakuraiG83, ShinS98} & Ours &
    Front & Top & Side & \cite{SakuraiG83, ShinS98} & Ours \tabularnewline
    \midrule

    \includegraphics[width=0.095\linewidth]{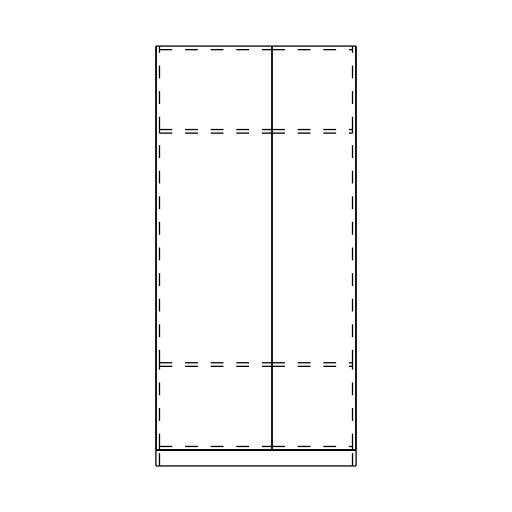} &
    \includegraphics[width=0.095\linewidth]{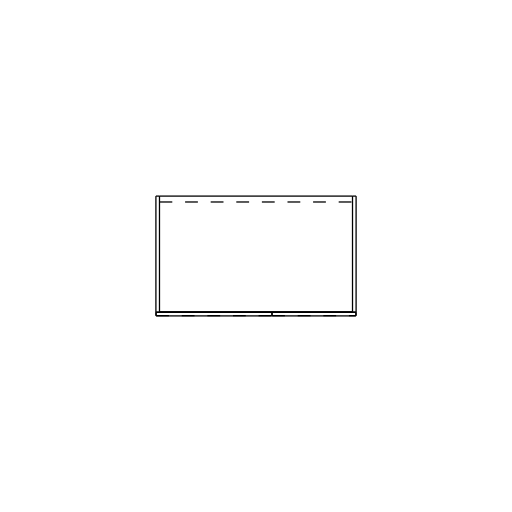} &
    \includegraphics[width=0.095\linewidth]{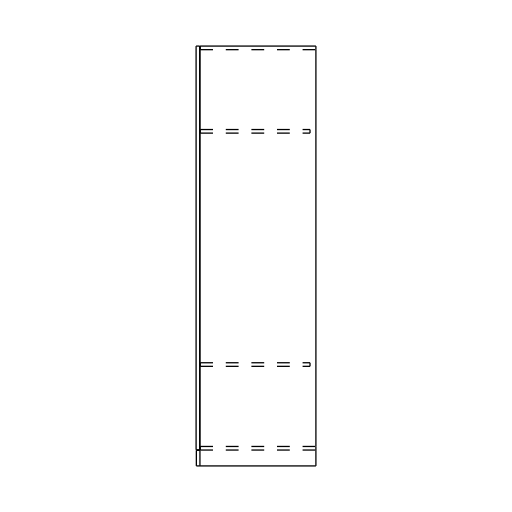} &
    \includegraphics[width=0.095\linewidth]{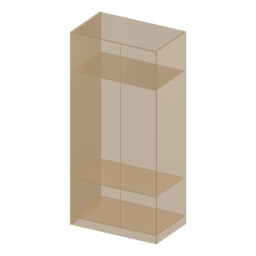} &
    \includegraphics[width=0.095\linewidth]{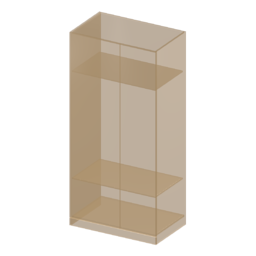} &

    \includegraphics[width=0.095\linewidth]{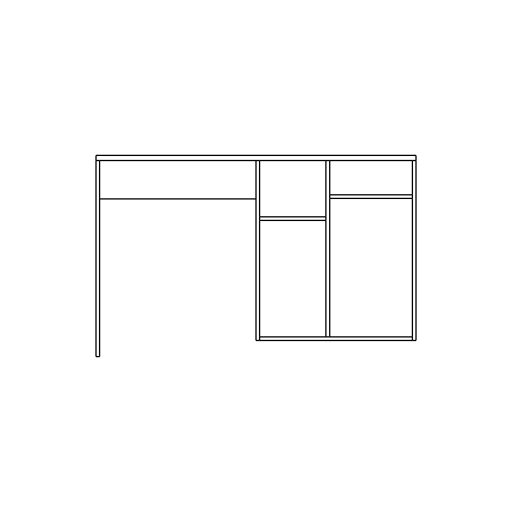} &
    \includegraphics[width=0.095\linewidth]{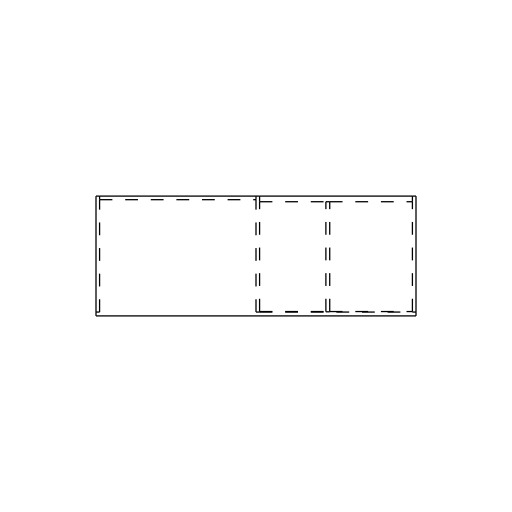} &
    \includegraphics[width=0.095\linewidth]{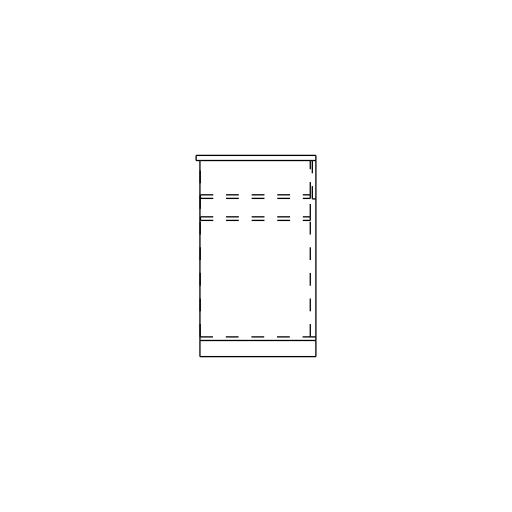} &
    \includegraphics[width=0.095\linewidth]{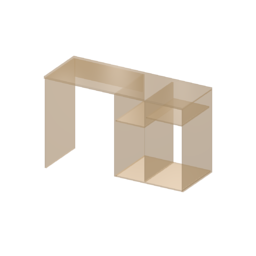} &
    \includegraphics[width=0.095\linewidth]{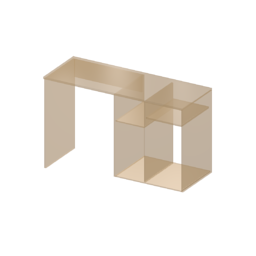}
    \tabularnewline

    \includegraphics[width=0.095\linewidth]{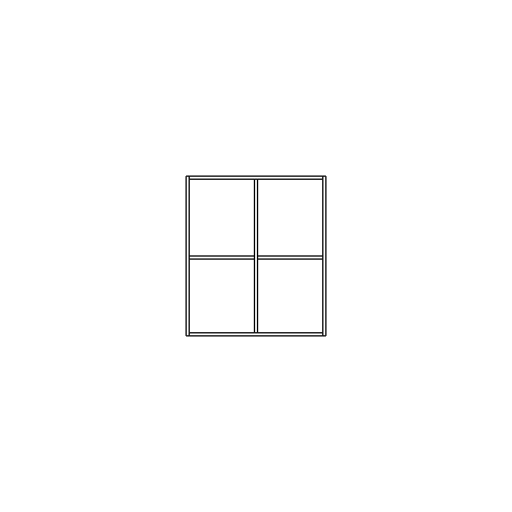} &
    \includegraphics[width=0.095\linewidth]{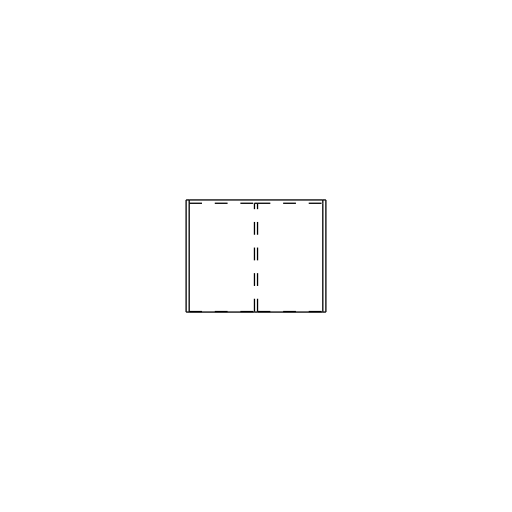} &
    \includegraphics[width=0.095\linewidth]{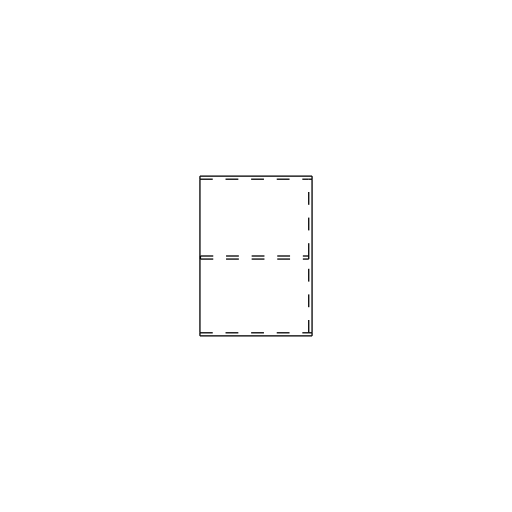} &
    {\color{red}\fbox{\includegraphics[width=0.095\linewidth]{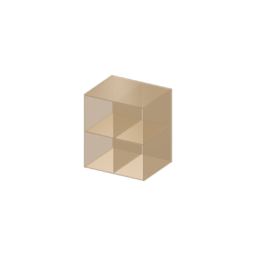}}} &
    \includegraphics[width=0.095\linewidth]{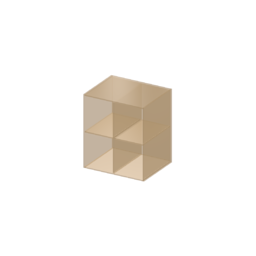} &

    \includegraphics[width=0.095\linewidth]{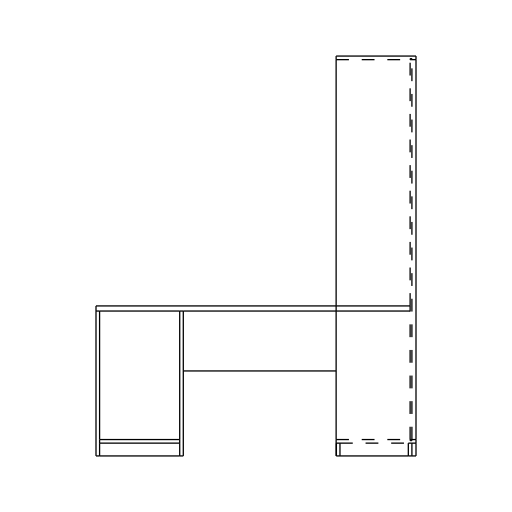} &
    \includegraphics[width=0.095\linewidth]{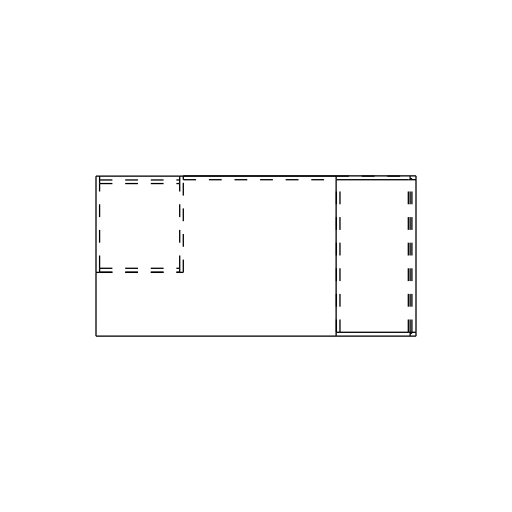} &
    \includegraphics[width=0.095\linewidth]{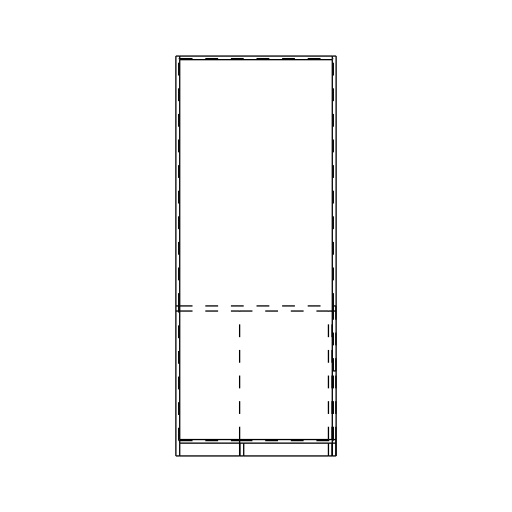} &
    \includegraphics[width=0.095\linewidth]{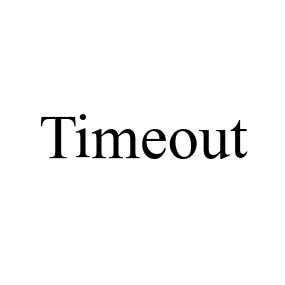} &
    \includegraphics[width=0.095\linewidth]{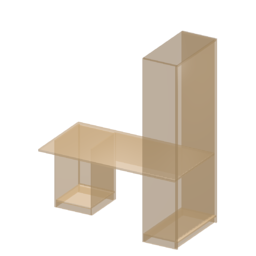}
    \tabularnewline

    \includegraphics[width=0.095\linewidth]{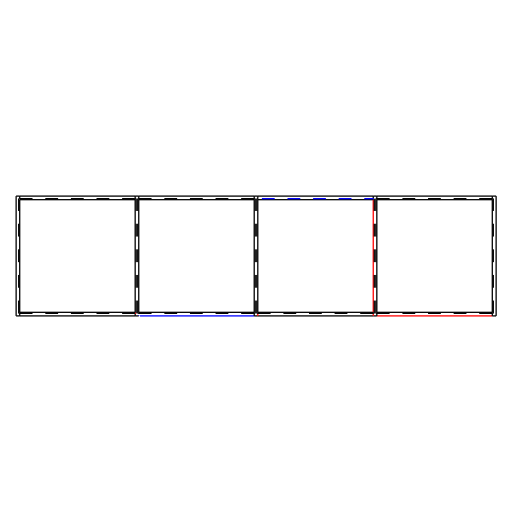} &
    \includegraphics[width=0.095\linewidth]{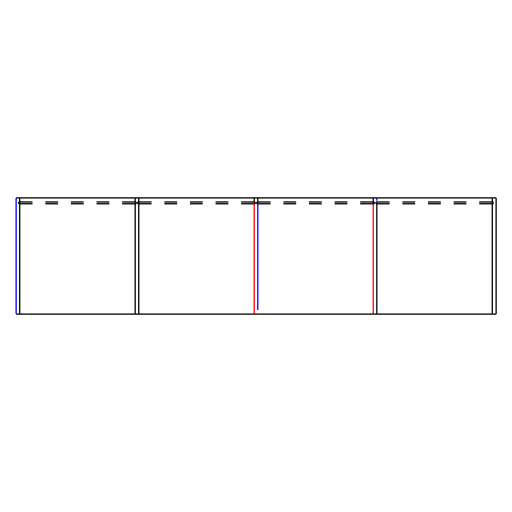} &
    \includegraphics[width=0.095\linewidth]{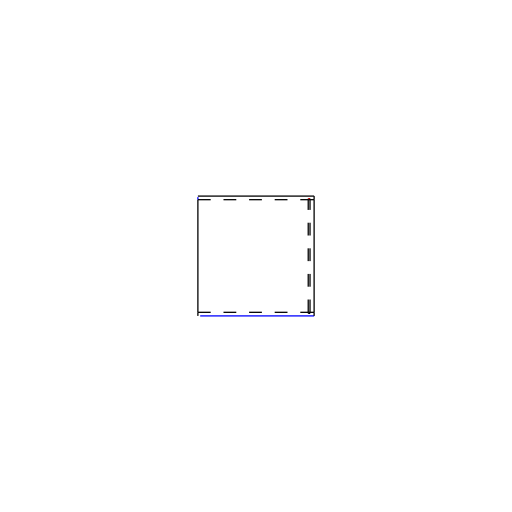} &
    {\color{red}\fbox{\includegraphics[width=0.095\linewidth]{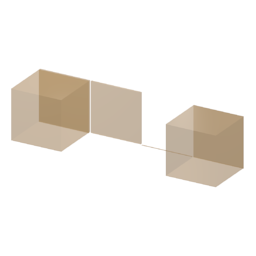}}} &
    \includegraphics[width=0.095\linewidth]{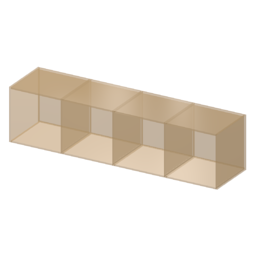} &

    \includegraphics[width=0.095\linewidth]{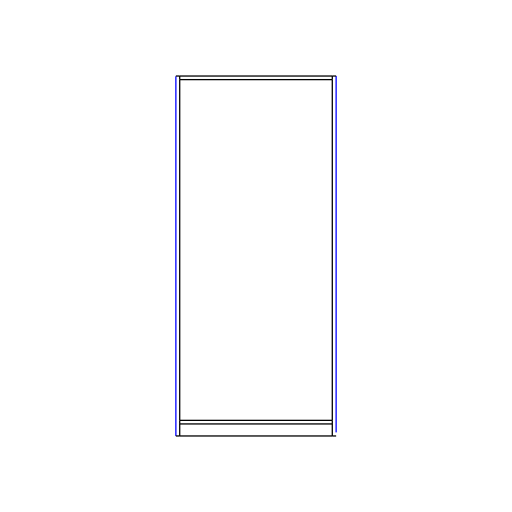} &
    \includegraphics[width=0.095\linewidth]{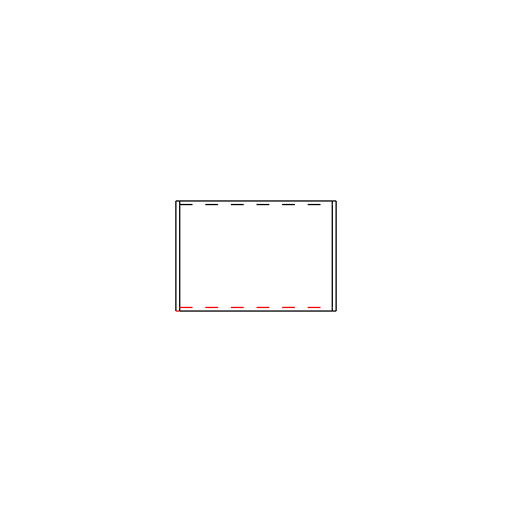} &
    \includegraphics[width=0.095\linewidth]{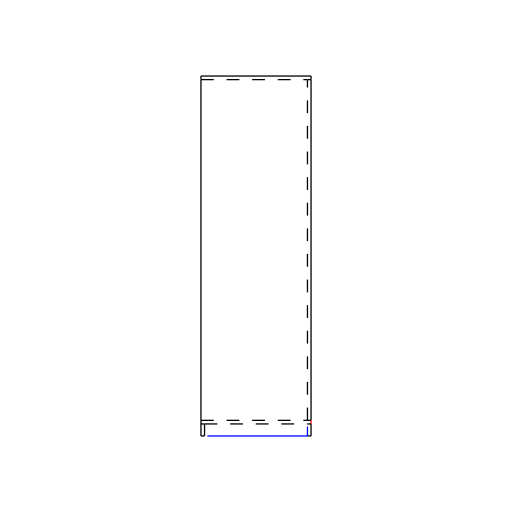} &
    {\color{red}\fbox{\includegraphics[width=0.095\linewidth]{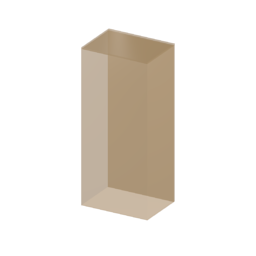}}} &
    \includegraphics[width=0.095\linewidth] {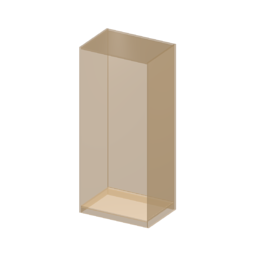} 
    \tabularnewline

    \includegraphics[width=0.095\linewidth]{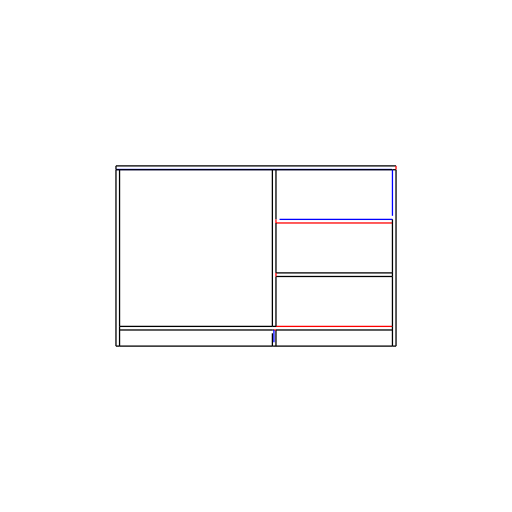} &
    \includegraphics[width=0.095\linewidth]{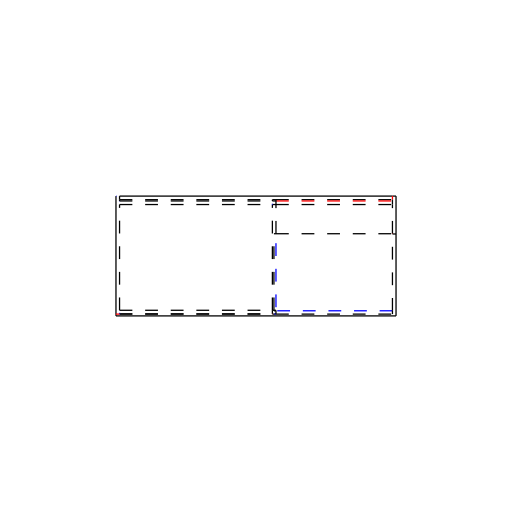} &
    \includegraphics[width=0.095\linewidth]{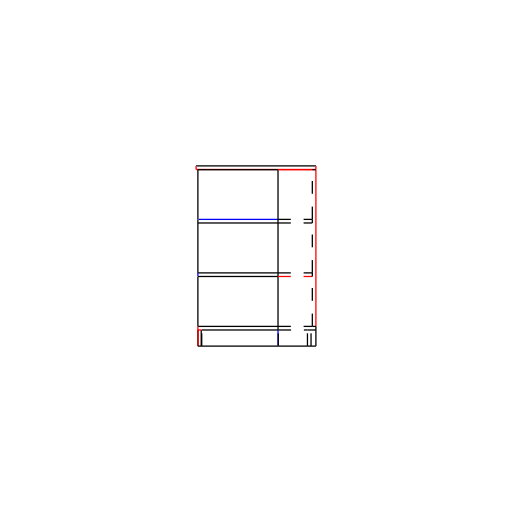} &
    {\color{red}\fbox{\includegraphics[width=0.095\linewidth]{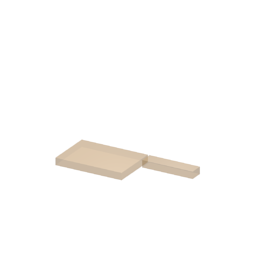}}} &
    \includegraphics[width=0.095\linewidth]{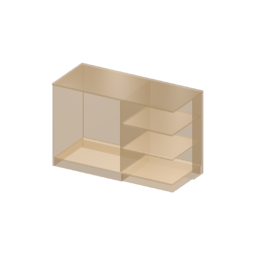} &

    \includegraphics[width=0.095\linewidth]{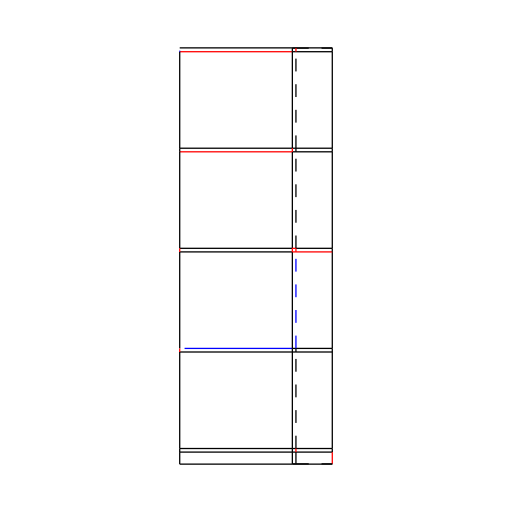} &
    \includegraphics[width=0.095\linewidth]{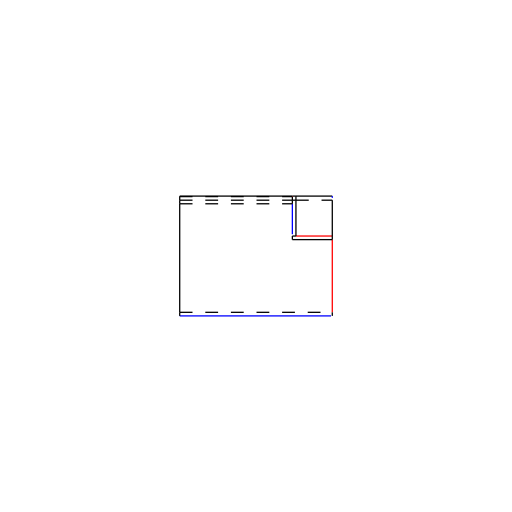} &
    \includegraphics[width=0.095\linewidth]{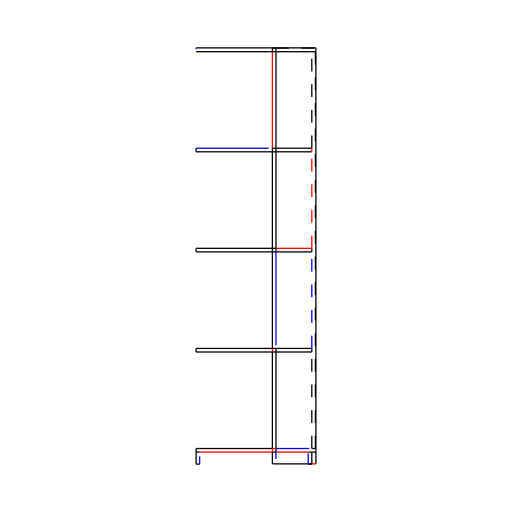} &
    \includegraphics[width=0.095\linewidth]{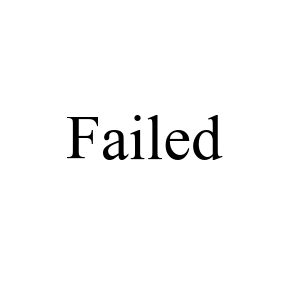} &
    \includegraphics[width=0.095\linewidth] {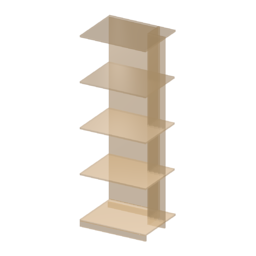}
    \tabularnewline

    \includegraphics[width=0.095\linewidth]{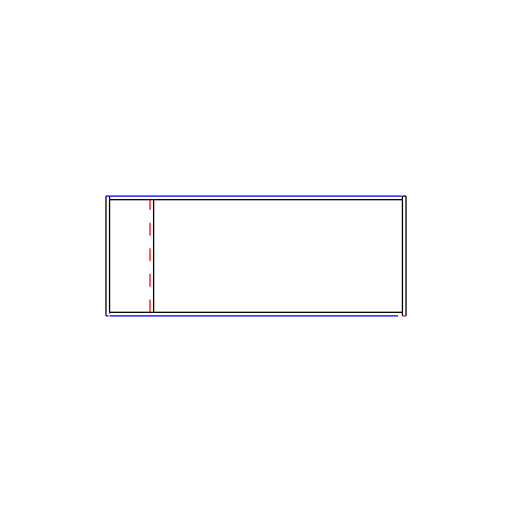} &
    \includegraphics[width=0.095\linewidth]{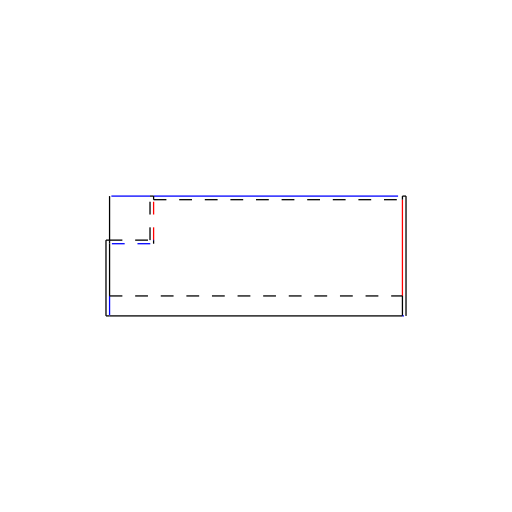} &
    \includegraphics[width=0.095\linewidth]{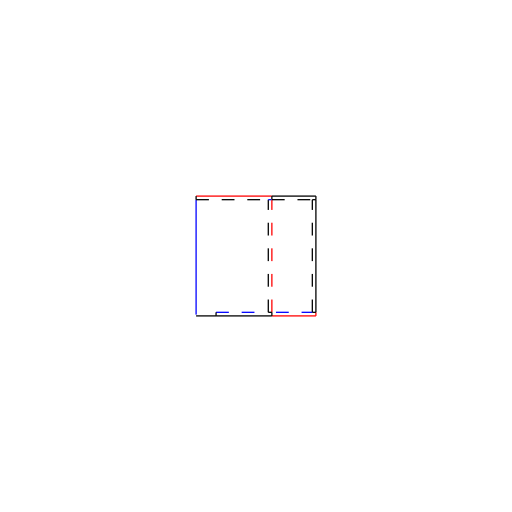} &
    \includegraphics[width=0.095\linewidth]{figures/failed.png} &
    \includegraphics[width=0.095\linewidth]{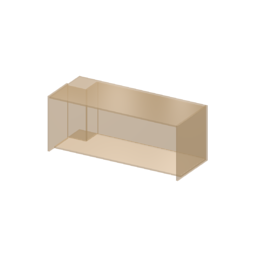} &

    \includegraphics[width=0.095\linewidth]{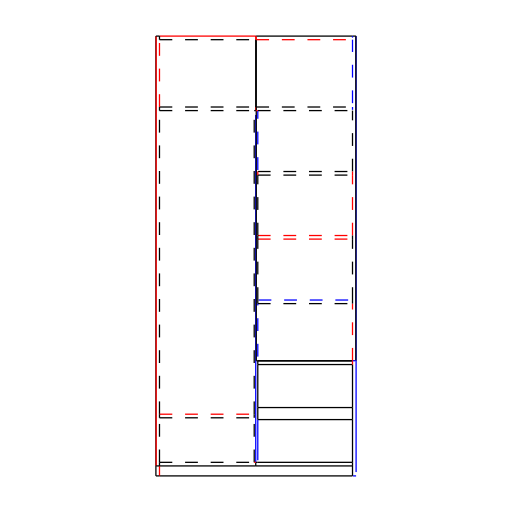} &
    \includegraphics[width=0.095\linewidth]{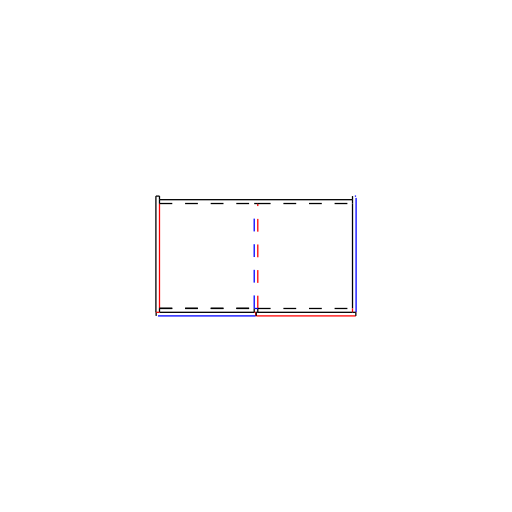} &
    \includegraphics[width=0.095\linewidth]{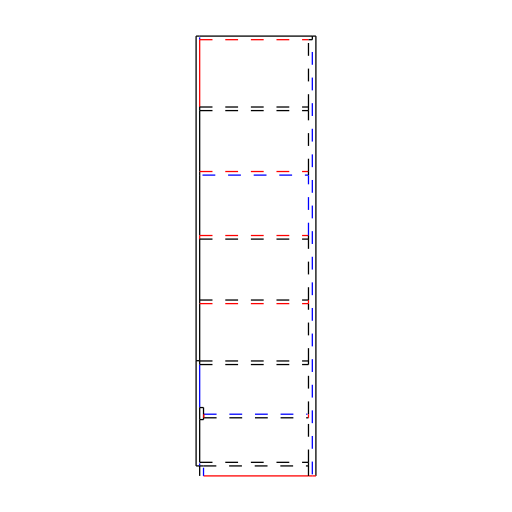} &
    \includegraphics[width=0.095\linewidth]{figures/failed.png} &
    \includegraphics[width=0.095\linewidth] {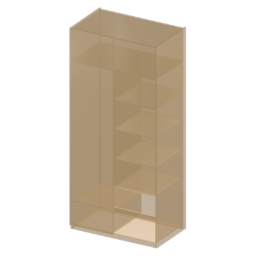}
    \tabularnewline

    \includegraphics[width=0.095\linewidth]{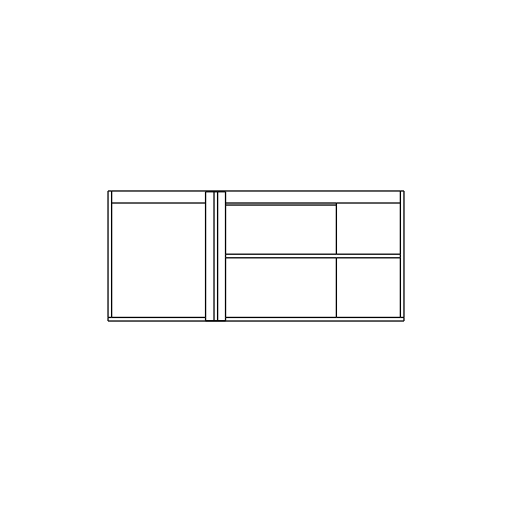} &
    \includegraphics[width=0.095\linewidth]{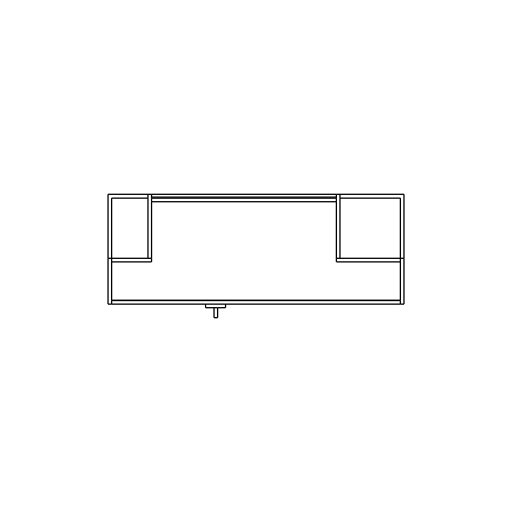} &
    \includegraphics[width=0.095\linewidth]{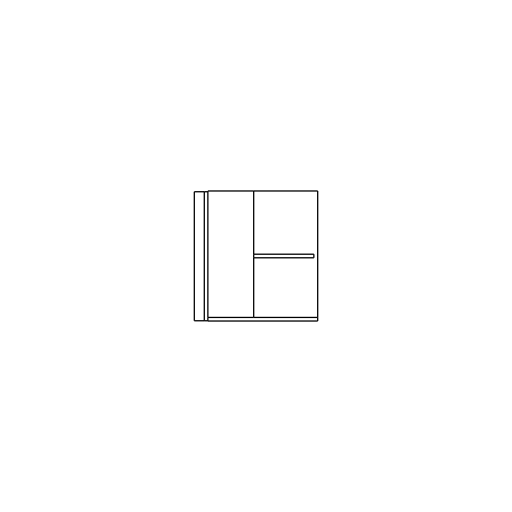} &
    \includegraphics[width=0.095\linewidth]{figures/failed.png} &
    \includegraphics[width=0.095\linewidth]{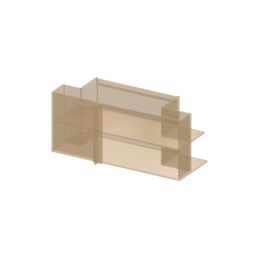} &

    \includegraphics[width=0.095\linewidth]{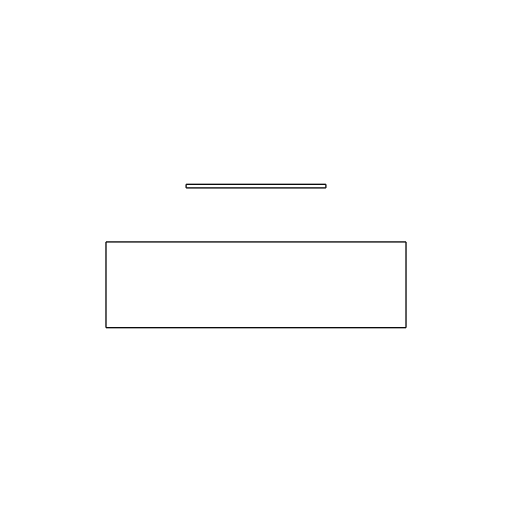} &
    \includegraphics[width=0.095\linewidth]{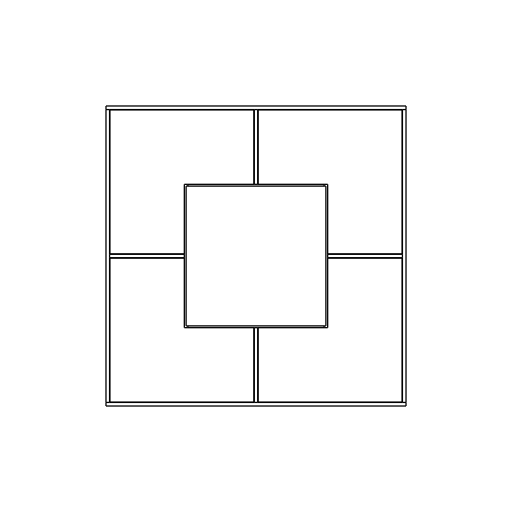} &
    \includegraphics[width=0.095\linewidth]{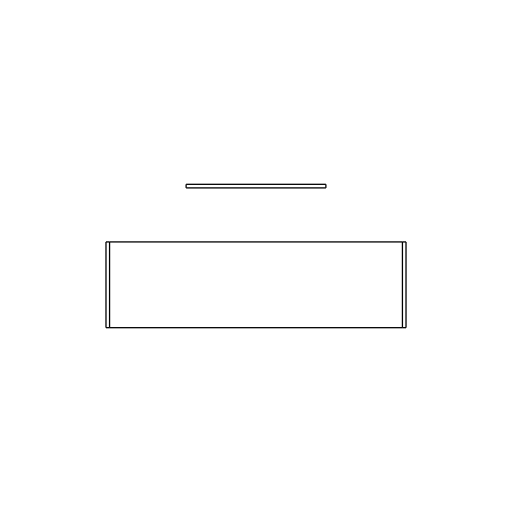} &
    {\color{red}\fbox{\includegraphics[width=0.095\linewidth]{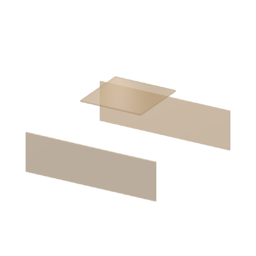}}} &
    \includegraphics[width=0.095\linewidth] {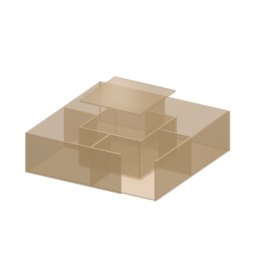}
    \tabularnewline

    \bottomrule
  \end{tabular}
  \caption{Qualitative results. {\bf Rows 1-2}: clean inputs. {\bf Rows 3-5}: Noisy inputs with noise level $10\%$, $20\%$, and $30\%$, respectively. {\bf Row 6}: Inputs with visible parts only. We use red boxes to indicate incorrect reconstructions.}\vspace{-3mm}
  \label{fig:tradition}
\end{figure*}

\smallskip
\noindent{\bf Qualitative results.} \cref{fig:tradition} visualizes some 3D reconstruction results of the two methods. In the \textit{first and second rows}, we show results with clean inputs. Our method correctly reconstructs all four objects, whereas the traditional pipeline fails on the last two objects. Specifically, for the first object in the second row, the traditional pipeline produces multiple solutions, and an incorrect one is selected as the final output. And for the second object, it fails to produce any result within the time budget (5 minutes).

In the \emph{third to fifth rows}, we show results from inputs with noise levels $10\%$, $20\%$, and $30\%$, respectively. As one can see, the performance of traditional pipeline degrades quickly. In particular, it fails to find any valid blocks for the two cases with noise level $30\%$. In contrast, our method correctly reconstructs all six objects in the three rows.

Finally, in the \emph{sixth row}, we show two cases from inputs with visible edges only. Again, the traditional approach performs poorly in these cases, whereas our method correctly recovers both objects.

\begin{figure}[t]
  \centering
  \includegraphics[width=0.95\linewidth]{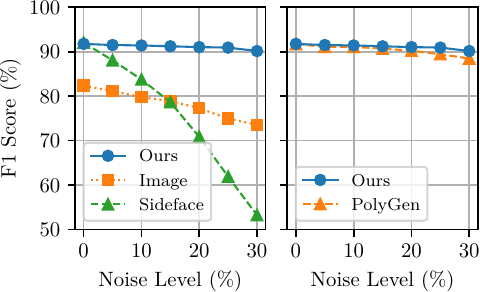}
  \caption{Ablation studies on the input sequence {\bf(left)} and the output sequence {\bf(right)}.} 
  \label{fig:ablations}
\end{figure}

\subsection{Ablation Studies}

Next, we investigate the effect of several design choices we made in the PlankAssembly model.

\smallskip
\noindent \textbf{Ablation study on the input.} First, we study the performance of our model with different types of inputs. In PlankAssembly, we directly use the sequence of 2D edges as input. Here, we consider two alternatives:

\noindent\underline{\em Image}: Many deep networks for 3D reconstruction use raster images as input. Inspired by Atlas~\cite{MurezABSBR20}, we replace the Transformer encoder in PlankAssembly with a CNN-based feature extractor to construct a 3D feature volume from posed images. Specifically, we use ResNet50-FPN~\cite{LinDGHHB17} to extract 2D features from each view. Then, we aggregate the 2D features into a 3D feature volume with known poses and use a 3D CNN to refine the 3D features. The Transformer decoder takes the flattened features as input and outputs the shape program.

\noindent\underline{\em Sideface}: Given the vectorized 2D line drawings, it is also possible to extract 2D sidefaces (\ie, rectangles that correspond to the 3D sidefaces of the plank models) and use the sequence of sidefaces as input. Intuitively, this allows the seq2seq model to leverage explicit correspondences between the input and output. To this end, we design a set of heuristic rules to extract sidefaces in each view: First, we use the \texttt{polygonize} API from Shapely~\cite{shapely} to construct minimal closed polygons from each line drawing. Then, each sideface is represented by a polygon's axis-aligned bounding box (AABB). Here, we exclude AABBs whose short side is larger than $\epsilon$ (those AABBs typically correspond to the profile faces of the planks). We also merge sidefaces recurrently if their short sides are adjacent. We set $\epsilon=50$mm in our experiments.

\cref{fig:ablations}~(left) shows the performance of our method \wrt different types of inputs. As one can see, using raster images as inputs results in lower F1 scores across all noise levels, possibly because the features extracted from the line drawings are very sparse when treated as images. Further, the model achieves similar accuracies on clean inputs when using lines or sidefaces. However, the performance of the model using sidefaces is more sensitive to noises in the input views. This again shows that attempts to establish explicit correspondences between input and output hurt the methods' robustness -- a phenomenon already seen in the comparison to traditional methods.

\smallskip
\noindent\textbf{Ablation study on the output sequence.} In PlankAssembly, we use shape programs as the outputs. In this experiment, we compare this choice to PolyGen~\cite{NashGEB20}, a popular approach to generate geometric models in the form of $n$-gon meshes. Similar to our method, PolyGen adopts a Transformer-based architecture and proceeds by generating a set of 3D vertices, which are then connected to form 3D faces. To obtain the planks in the cabinet models, we borrow the block generation step in the traditional pipeline to construct closed solids from the predicted faces.

\begin{figure}[t]
  \centering
  \setlength{\tabcolsep}{2pt}
  \setlength{\fboxsep}{0pt}
  \begin{tabular}{c|c|c}
    \toprule
    PolyGen (face) & PolyGen (solid) & Ours \tabularnewline
    \midrule
    {\color{red}\fbox{\includegraphics[width=0.30\linewidth]{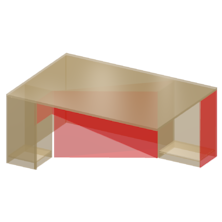}}} &
    {\color{red}\fbox{\includegraphics[width=0.30\linewidth]{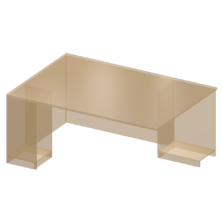}}} &
    \includegraphics[width=0.30\linewidth]{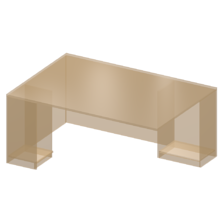}
    \tabularnewline
    {\color{red}\fbox{\includegraphics[width=0.30\linewidth]{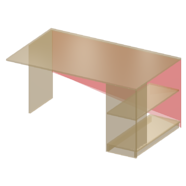}}} &
    {\color{red}\fbox{\includegraphics[width=0.30\linewidth]{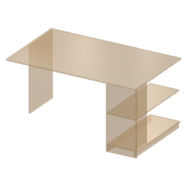}}} &
    \includegraphics[width=0.30\linewidth]{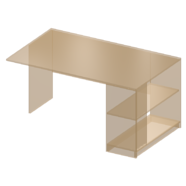}
    \tabularnewline
    \bottomrule
  \end{tabular}
  \caption{Qualitative comparison with PolyGen~\cite{NashGEB20}. Input views are omitted. For PolyGen, we show results both before and after the solid construction step. Some incorrectly reconstructed faces are highlighted in red.}
  \label{fig:polygen}
\end{figure}

The results are shown in \cref{fig:ablations}~(right). Our approach outperforms PolyGen, especially with high noise levels. Besides, our method runs about six times faster than PolyGen ($0.63$ \vs $3.61$ seconds per sample), partly because PlankAssembly directly generates planks as the output and has a shorter output sequence (solids \vs vertices$+$faces).

\cref{fig:polygen} compares the 3D models generated by our method and PolyGen. One notable issue with PolyGen is that since it generates each face separately, there is no guarantee that the faces will form closed solids (\ie, planks). For example, in the second row of \cref{fig:polygen}, one face predicted by PolyGen has a non-rectangular shape, leading to missing planks after the solid construction step.

\begin{figure}[t]
  \centering
  \setlength{\tabcolsep}{1pt}
  \begin{tabular}{c|c|c|c}
    \includegraphics[width=0.24\linewidth]{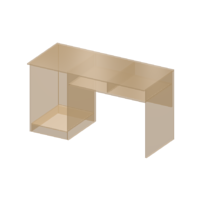} &
    \includegraphics[width=0.24\linewidth]{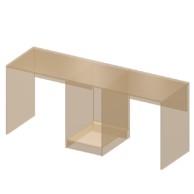} &
    \includegraphics[width=0.24\linewidth]{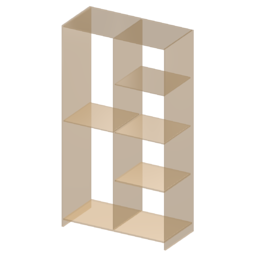} &
    \includegraphics[width=0.24\linewidth]{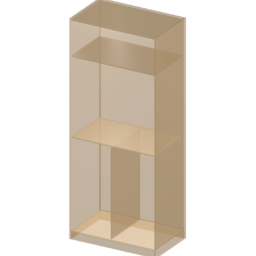}
    \tabularnewline
    \includegraphics[width=0.24\linewidth]{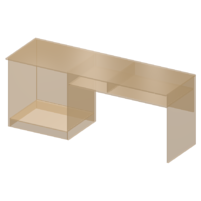} &
    \includegraphics[width=0.24\linewidth]{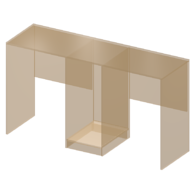} &
    \includegraphics[width=0.24\linewidth]{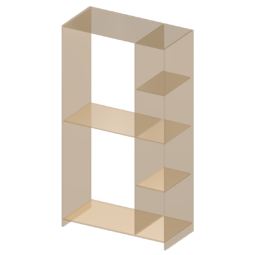} &
    \includegraphics[width=0.24\linewidth]{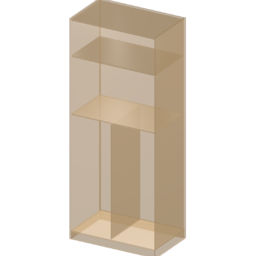}
    \tabularnewline
    (a) & (b) & (c) & (d)
    \tabularnewline
  \end{tabular}
  \caption{Example of simple user edits. {\bf Top:} models reconstructed by PlankAssembly. {\bf Bottom:} edited models.}
  \label{fig:editing}
\end{figure}

Another benefit of leveraging domain-specific language over general geometric forms such as $n$-gons is that the generated shapes can better support user edits in the CAD modeling software. Specifically, given the attachment relationships predicted by our method, a cabinet model may undergo global scaling operations (\cref{fig:editing}~(a-b)) or local editing operations (\cref{fig:editing}~(c-d)) while maintaining the correct topology.

\subsection{Failure Cases}

\begin{figure}[t]
  \centering
  \setlength{\tabcolsep}{1pt}
  \begin{tabular}{c|c|c|c|c}
    \toprule
    Front & Top & Side & Ours & GT
    \tabularnewline
    \midrule
    \includegraphics[width=0.19\linewidth]{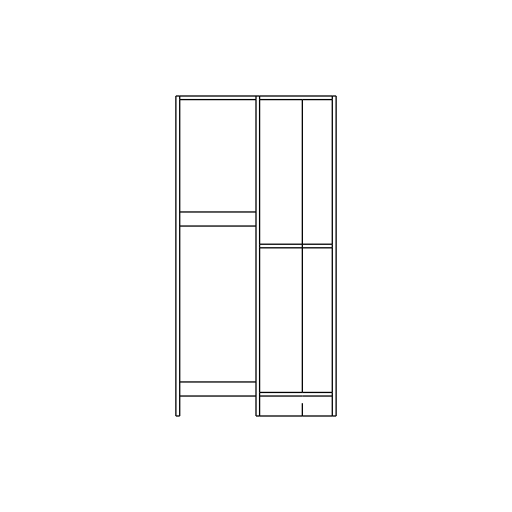} &
    \includegraphics[width=0.19\linewidth]{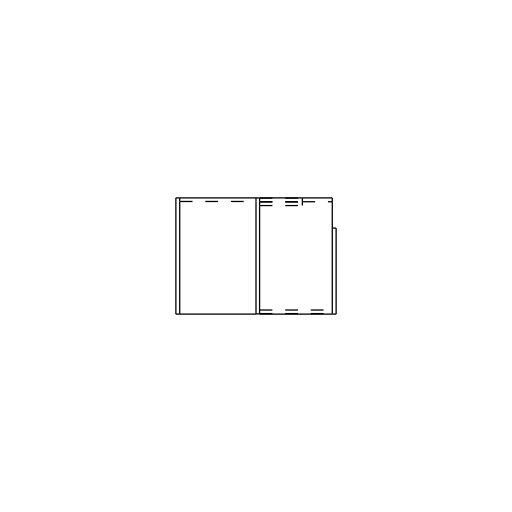} &
    \includegraphics[width=0.19\linewidth]{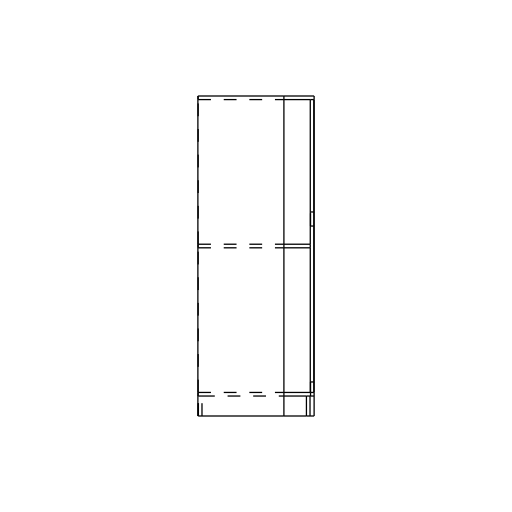} &
    \includegraphics[width=0.19\linewidth]{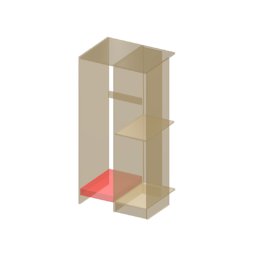} &
    \includegraphics[width=0.19\linewidth]{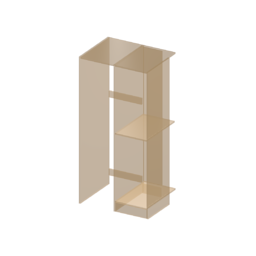}
    \tabularnewline
    \includegraphics[width=0.19\linewidth]{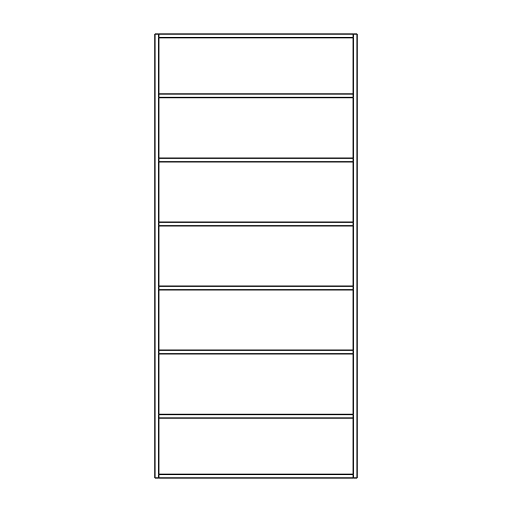} &
    \includegraphics[width=0.19\linewidth]{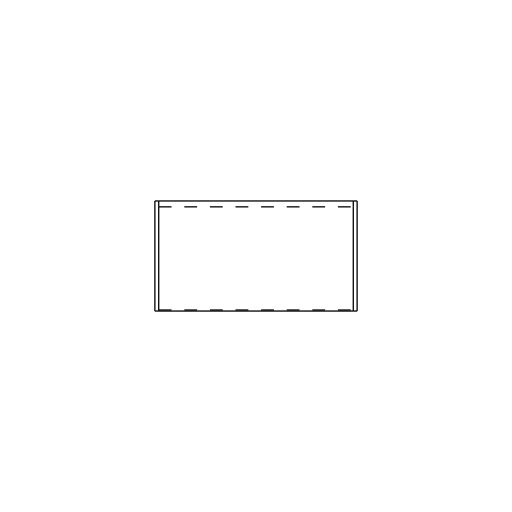} &
    \includegraphics[width=0.19\linewidth]{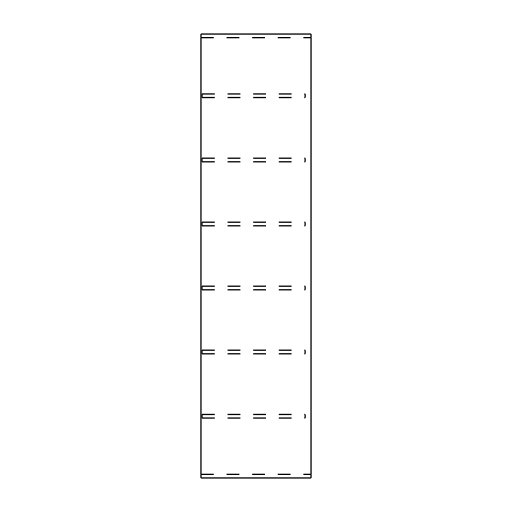} &
    \includegraphics[width=0.19\linewidth]{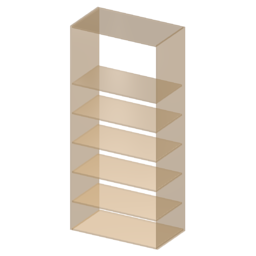} &
    \includegraphics[width=0.19\linewidth]{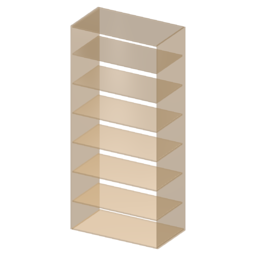}
    \tabularnewline
    \bottomrule
  \end{tabular}
  \caption{Failure cases. We highlight incorrectly reconstructed planks in red.}
  \label{fig:failure}
\end{figure}

\cref{fig:failure} illustrates two most common failure modes of our PlankAssembly model. In the first example, the network makes incorrect predictions for attachments. In the second example, the reconstructed 3D model is incomplete because the stop token is predicted too early, which is a known issue with auto-regressive models.

\section{Discussion}
\label{sec:discussion}

This paper advocates a \emph{generative approach} to 3D CAD model reconstruction from three orthographic views. Two lessons can be learned from our experiments: \emph{First}, compared to finding explicit correspondences between the 2D line drawings and 3D models, the attention mechanism plays a key role in the deep network's robustness to the imperfect inputs. \emph{Second}, incorporating domain knowledge in the generative model benefits both the reconstruction and downstream applications.

One may argue that our experiments are limited to cabinet furniture, a special type of CAD model. However, we emphasize that our main idea and the lessons learned are general and can be applied to any CAD model. For example, prior work such as DeepCAD~\cite{WuXZ21} has developed neural networks which are able to generate CAD command sequences suitable for mechanical parts. Unlike cabinet furniture, mechanical parts often have non-rectangular profiles (but fewer blocks). It is thus relatively straightforward to extend our approach to such domains.

A more challenging scenario is one attempting to apply our data-driven approach to domains where large-scale CAD data is unavailable or even nonexistent, such as buildings or complex mechanical equipment. Besides, our current approach does not consider other information available in CAD drawings, such as layers, text, symbols, and annotations. Recently, several methods have been proposed for panoptic symbol spotting in CAD drawings~\cite{FanZLCZT21, ZhengLZLPT22, FanCWW22}. We believe that such information is also vital for 3D reconstruction from complex CAD drawings.

\section*{Acknowledgements}

This work was supported in part by the Key R\&D Program of Zhejiang Province (2022C01025). Jian Yin is supported by the National Natural Science Foundation of China (U1911203, U2001211, U22B2060), Guangdong Basic and Applied Basic Research Foundation (2019B1515130001), Key-Area Research and Development Program of Guangdong Province (2020B0101100001).

{\small
\bibliographystyle{ieee_fullname}
\bibliography{bibliography}
}

\end{document}